\documentclass[a4paper,fleqn]{cas-dc}

\usepackage[numbers]{natbib}

\usepackage{graphicx}
\usepackage{times}
\usepackage{epsfig}
\usepackage{graphicx}
\usepackage{amsmath}
\usepackage{amssymb}
\usepackage{bm}
\usepackage{multirow}

\usepackage{caption}
\usepackage{subcaption}
\usepackage{bbding}
\usepackage{color}
\usepackage{hyperref}

\usepackage{amsfonts}
\usepackage{tabularx}

\def\tsc#1{\csdef{#1}{\textsc{\lowercase{#1}}\xspace}}
\tsc{WGM}
\tsc{QE}

\begin{document}
\let\WriteBookmarks\relax
\def\floatpagepagefraction{1}
\def\textpagefraction{.001}

\shorttitle{Cross-source Point Cloud Registration: Challenges, Progress and Prospects}    

\shortauthors{Huang et al.}  

\title [mode = title]{Cross-source Point Cloud Registration: Challenges, Progress and Prospects}  



%

\author[a]{Xiaoshui Huang} 






\affiliation[a]{organization={Shanghai AI Laboratory, China},
            city={Shanghai},
            country={China}}

\author[b]{Guofeng Mei}
\author[b]{Jian Zhang}
\fnmark[*]

\ead{Jian.Zhang@uts.edu.au}



\affiliation[b]{organization={GBDTC, FEIT, University of Technology Sydney}}

\cortext[1]{Corresponding author}


\begin{abstract}
The emerging topic of cross-source point cloud (CSPC) registration has attracted increasing attention with the fast development background of 3D sensor technologies. Different from the conventional same-source point clouds that focus on data from same kind of 3D sensor (e.g., Kinect), CSPCs come from different kinds of 3D sensors (e.g., Kinect and { LiDAR}). CSPC registration generalizes the requirement of data acquisition from same-source to different sources, which leads to generalized applications and combines the advantages of multiple sensors. In this paper, we provide a systematic review on CSPC registration. We first present the characteristics of CSPC, and then summarize the key challenges in this research area, followed by the corresponding research progress consisting of the most recent and representative developments on this topic. Finally, we discuss the important research directions in this vibrant area and explain the role in several application fields.
\end{abstract}




\maketitle


\begin{abstract}
	The emerging topic of cross-source point cloud (CSPC) registration has attracted increasing attention with the fast development background of 3D sensor technologies. Different from the conventional same-source point clouds that focus on data from same kind of 3D sensor (e.g., Kinect), CSPCs come from different kinds of 3D sensors (e.g., Kinect and { LiDAR}). CSPC registration generalizes the requirement of data acquisition from same-source to different sources, which leads to generalized applications. In this paper, we provide a systematic review on CSPC registration. We first present the characteristics of CSPC, and then summarize the key challenges in this research area, followed by the corresponding research progress consisting of the most recent and representative developments on this topic. Finally, we discuss the important research directions in this vibrant area.
\end{abstract}

\section{Introduction}
Point cloud is a collection of 3D points that depicts the surface of a 3D scene. Recently, point cloud acquisition becomes widely available and consumer affordable because of fast development of sensor technologies. Because of the complexity of 3D data acquisition, there is no perfect sensor for 3D data acquisition. The existing sensors have specific advantages and limitations in recording the 3D scenes. For example, { LiDAR} utilizes the active light to capture accurate but sparse point clouds. Depth camera uses infrared or stereo vision technologies to estimate depth and the depth could be utilized to generate dense point clouds but usually with limited range and the accuracy is moderate. RGB camera combined with 3D reconstruction technologies can also generate dense point cloud with texture information but the accuracy is usually lower than depth camera and { LiDAR} sensor. Because of coexist of advantages and limitations, fusion of different kinds of sensor data can combine the advantages of many sensors so that the point cloud acquisition could be accurately, efficiently and detailed enough. Among this fusion process, CSPC registration technology plays the critical role.

\begin{figure}
	\includegraphics[width=\linewidth]{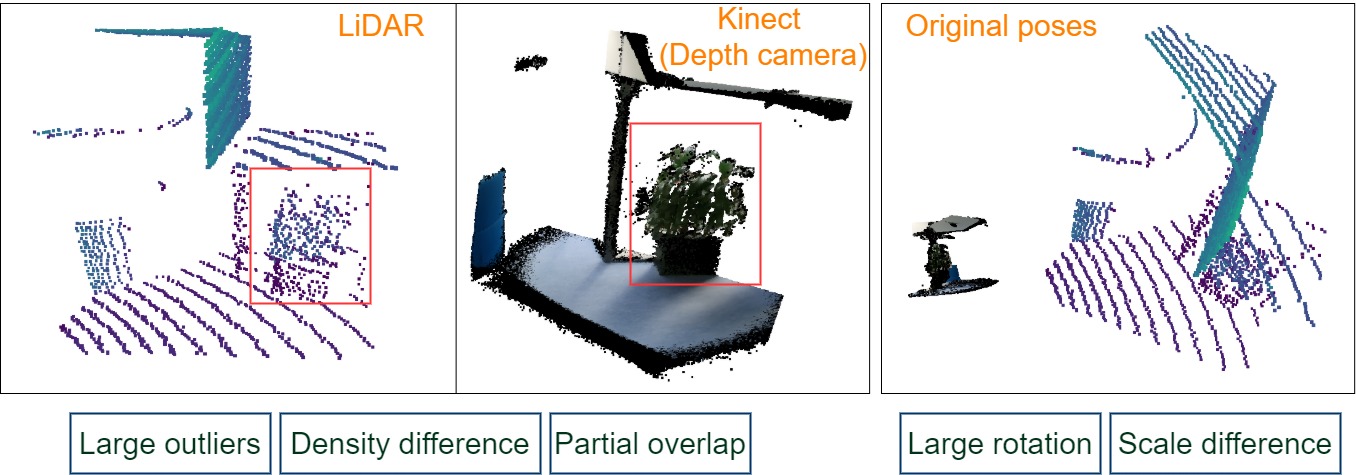}
	\caption{An example to show the challenges in the cross-source point clouds.}
	\label{f1}
\end{figure}

Excepting the aforementioned sensor fusion, cross-source is a important computer vision technology because of its critical role in many field applications. Firstly, \textbf{robotics}. The high accurate sensors provide high-quality data from the server side. Then, the reconstruction and localization service can be achieved by cheap and consumer-available sensor data from user side. This would deploy the service in a low-cost way from user side. Secondly, \textbf{construction}. Comparing the cross-source of designed 3D CAD models and point cloud scans on the site can provide much quick evaluation for the construction quality. Thirdly, \textbf{remote sensing}. Cross-source point clouds can merge different level of details for remote sensing and merge different sensors to leverage different merits for remote sensing.

The CSPC registration is a technology to align two point clouds from different kinds of sensors. Its scientific problem lies in the transformation matrix (\emph{scale}, rotation and translation) estimation of two overlapped point clouds from different types of sensors. From the view of scientific problem, CSPC registration needs to estimate the scale difference which is different to the previous same-source rigid point cloud registration.

The CSPC registration is extremely challenging because the handled point clouds from different types of sensors contain numerous variations. According to \cite{peng2014street,huang2017systematic,huang2017coarse,mellado2015relative}, these variations are defined as cross-source challenges that include 1) large outliers, 2) density difference, 3) partial overlap, 4) large rotation, 5) scale difference. The cross-source challenges are difficult since these variations are usually mixture and significant in the cross-source point clouds. Figure \ref{f1} visually shows an example of the challenges.

The cross-source challenges have been overlooked. Most of the existing techniques give meaningful results only if the input data are of the same type and density (resolution), and only contains rigid transformation. As a result, many recent state-of-the-art registration methods face difficulty or even fail when meeting with cross-source challenges.  For example, DGR \cite{choy2020deep} achieves 91.3\% at 3DMatch \cite{zeng20173dmatch} (same-source dataset captured by depth camera) but only 36.6\% at 3DCSR (cross-source dataset) \cite{huang2021comprehensive}. FMR \cite{huang2020feature} achieves less than 2$^\circ$ at 7Scene (a same-source dataset capture by depth camera) but only 17.8\% at 3DCSR. The main reasons of this big performance gap are two aspects. Firstly, these variations are usually significant and mixture so that both the robust feature extraction and correspondence search {  face difficulty}. This will result in large amount of correspondence outliers in the correspondence-based algorithms. Secondly, since the mixture variations make the robust feature extraction difficult, the correspondence-free methods also face challenge.

The CSPC registration deserves more attention because of two reasons. Firstly, as discussed above, point cloud acquired from single sensor contains limitations and CSPC registration merges multiple sensors' data which could overcome the limitations of single sensor. This largely contributes the data acquisition for the AI community. Secondly, CSPC registration plays undeniable and critical role in revolutionizing many downstream tasks in the AI community, such as 3D reconstruction and localization.  1) 3D reconstruction. { 3D reconstruction by leveraging multiple sensors provides accurate}, detailed and efficient 3D environment map construction for many applications like autonomous driving, robotics and metaverse. For example, LiDAR provides a quick and accurate way to generate large map but usually sparse. Combined with other 3D sensors such as depth camera could generate accurate and detailed maps. 2) Localization. Point cloud can provide more accurate visual localization service. However, the service providers usually adopt expensive 3D sensors to guarantee the service quality while the consumers usually use cheap 3D sensors to consider the cost. The CSPC registration is the indispensable technique to guarantee the localization accuracy. Although the CSPC registration is a cornerstone technology in the AI community, the development is much lacked behind compared to the previous same-source point cloud registration. With the fast development of sensor technology, it is the right time and urgent to conduct research on this emerging topic. However, there is no systematical review about CSPC registration to clearly expose the value of this research and summarize the challenges.

In this paper, we provide a comprehensive survey about CSPC registration aiming to clearly expose the value of CSPC registration and provide insights to the further research. We first present the characteristics of CSPC, and deeply analyze the key challenges in this research area. Then, we present the corresponding research progress consisting of the most recent and representative developments on this topic. After that, we discuss the important research directions in this vibrant area.  Finally, we summarize the potential application fields of cross-source point clouds registration and explain the valuable role of CSPC registration. The contributions could be summarized as below:

\begin{itemize}
	\item Summarize the characters of the existing 3D sensors and analyze the cross-source challenges.
	\item A comprehensive literature review about CSPC registration.
	\item Propose several the research directions based on our literature analysis.
	\item Summarize the application fields and explain the role of CSPC registration.
\end{itemize}

\section{3D sensors and cross-source challenges}
Recent several years, the sensor technologies have endured fast development. Many 3D sensors available and several of them are consumer affordable. In this paper, we list the recent available sensor types that could produce point clouds and summarize their characterizes.  

\begin{table}[h]
	\begin{center}
		\begin{tabular}{p{1.5cm}|p{1.8cm}|c}\hline
			Sensor & Mechanism & Characterizes\\\hline
			LiDAR       & Light             & Sparse, 1-500 meter, XYZ \\ \hline
			Depth-A       & Infrared          & Dense,  0.5-9 meter, RGBD \\ \hline
			Depth-P       & stereo vision     & Dense,  0.2-20 meter, RGBD \\ \hline
			RGB-Cam  & 3D reconstruction & Dense,  ~ , XYZ-RGB \\ \hline
		\end{tabular}
	\end{center}
	\caption{The summary of recent sensor types that could generate point clouds.}
	\label{tsensor}
\end{table}

Table \ref{tsensor} shows that different sensor types have different imaging  mechanisms. This section will introduce the mechanisms, characterizes and compare their difference. 

\begin{itemize}
	\item \emph{LiDAR}: sends a brunch of light and receives the reflected light. The time difference between sender and receiver measures the distance between the sensor to the surface. {  Then, the distances combining the sensor's XY coordinate system generate the 3D points}. The LiDAR point clouds usually have a long range (up to 500 meter) while they are usually sparse and the data format is XYZ coordinates and covering 360$^\circ$ around the sensor. 
	
	\item \emph{Depth-A}: The active depth sensor (Depth-A)  sends infrared light and measures the depth by receiving it or recognizing the shape distortion. Then, the depth is converted into Z coordinate using camera interior parameters.  {  Finally, the point clouds are generated by combining the Z coordinate with the sensor's XY coordinates}. The acquired point clouds are usually dense and a small range (0.2-9 meter).  
	
	\item \emph{Depth-P}: The passive depth sensor (Depth-P) utilizes the stereo method to estimate the depths. Then, the point cloud is generated with the same way of Depth-A by using the depth. The acquired point clouds are usually dense and a moderate range (0.2-20 meter).

	\item \emph{RGB-Cam}: RGB camera (RGB-Cam) combines with the 3D reconstruction techniques \cite{zhou2014color} structure from motion (SFM) can also be utilized to generate the point clouds. The reconstructed point clouds are usually dense and colored.
\end{itemize}

\begin{table*}[ht]
	\centering
	\begin{tabularx}{0.9\textwidth}{l|X|X}\hline
		Cross-source sensors    & Existing solutions & Key process\\\hline
		LiDAR, RGB camera (SFM) & {\cite{peng2014street},\cite{swetnam2018considerations}, \cite{tazir2018cicp}, \cite{li2021linear},} \cite{li2021laplacian}, \cite{li2021point},\cite{huang2016coarse}  & correspondence optimization          \\ \hline
		LiDAR, Depth camera    &  \cite{ling2022graph}  & model matching  \\ \hline
		Depth camera, RGB camera (SFM) &  CSGM\cite{huang2017systematic},GCTR\cite{huang2019fast},\cite{huang2017coarse}, \cite{liu2021matching}, \cite{huang2020feature}   & model matching, feature learning            \\ \hline
	\end{tabularx}
	\caption{The summary of different kinds of sensors that used in the existing solutions.}
	\label{t1}
\end{table*} 

Different imaging mechanisms have different merits and limitations in point cloud acquisition. { LiDAR} is more suitable for long range and large scale scenes. Depth sensor is more suitable to close range scene acquisition with more details. RGB-cam + 3D reconstruction has slight long range than depth sensor but with slightly less details. The CSPC registration can combine merits of multiple sensors and overcome the limitation of single sensor. This is a highly valued research for many fields, such as remote sensing \cite{swetnam2018considerations}, robotics \cite{li2021point} and construction \cite{huang2017coarse}. { Table \ref{t1} summarizes the different kinds of sensors that used in the the existing CSPC registration solutions. From the Table \ref{t1} we can see that, the key process of LiDAR and RGB camera fusion methods is correspondence optimization, and the fusion of LiDAR and Depth camera use the key process of model matching. In comparison, the methods for depth camera and RGB camera fusion, the key process has wide choice options including model matching and feature learning. }

The generated cross-source point clouds usually contain large variations (see Figure \ref{f1} as a visual example). The registration of cross-source point clouds face much more challenges than the same-source point clouds (e.g., all point clouds are captured with LiDAR sensor). According to \cite{huang2017systematic}, the cross-source point clouds are mainly contains the following challenges:

\begin{itemize}
	\item \emph{Outliers}. Because of different imaging mechanisms and capture time, it is difficult to guarantee the same 3D points be located at same positions. Also, it is difficult to generate the same amount of 3D points in two acquired point clouds. Both these variations will result in acquisition outliers. Moreover, during the correspondence estimation, the inaccurate correspondences are another kind of outliers. 
	
	In the cross-source point clouds, the outliers are much difficult than the same-source. In same-source cases, the sensor accuracy is the same and the outliers come from different positions. In cross-source cases, the sensor accuracy is different and the outliers come from the different positions or sensing noisy. For example, the flowerpot in Figure \ref{f1} shows slightly different shapes. These outliers come from different sensor noise are much challenges than same-source outliers.
	
	\item \emph{Density difference}. Because of different imaging mechanisms and characterizes of different sensors, the acquired point clouds usually have different kinds of density. For example, LiDAR point clouds are sparse while stereo camera point clouds are dense.
	
	\item \emph{Partial overlap}. The captured point clouds from different kinds of sensors are difficult to guarantee the exactly the same poses and field of views. Therefore, cross-source point clouds are usually partially overlapped. The partial overlap problem is widely existed in the cross-source point clouds and need to pay more attention than same-source point clouds.
	
	\item \emph{Large rotation}. Different to the sequential point cloud acquisition, the CSPC acquisition faces difficult to keep the small rotation. In real scenes, the acquisition poses are randomly between two captured point clouds which may often contain large rotation.
	
	\item \emph{Scale difference}. Because of different imaging mechanisms, the physical mean of each unit in their captured point clouds are usually not consistent. Without any calibration, the cross-source point clouds contain scale difference. For example, one unit in { LiDAR} point cloud represent 1 meter but one unit in stereo camera may represent 20 meters.
\end{itemize}



\begin{figure*}[ht]
	\includegraphics[width=0.9\linewidth]{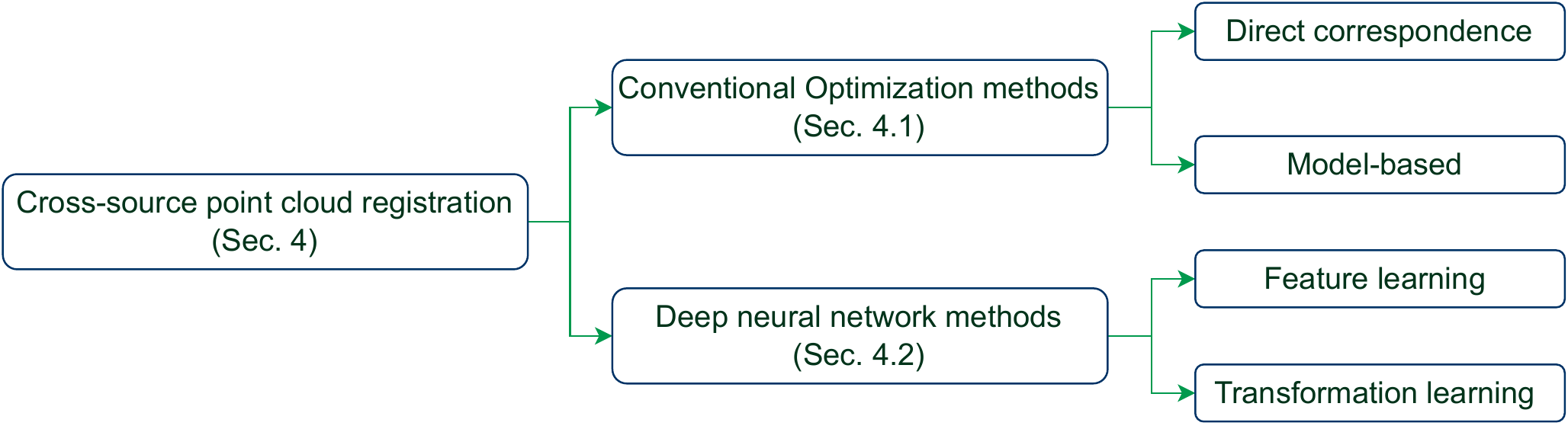}
	\caption{A categorization of CSPC registration from the technical perspective. }
	\label{category}
\end{figure*}
\section{Problem formulation}
The goal of CSPC registration is to estimate the transformation matrix (rotation $R\in \mathbb{R}^{3\times 3}$, translation $t \in \mathbb{R}^{1\times3}$ and scale $s$) that can align two point clouds from different types of sensors. Let $X\in \mathbb{R}^{N\times3}$ and $Y\in \mathbb{R}^{M\times 3}$ be two point clouds from two different types of sensors respectively. The problem of CSPC registration is formulated to optimize the objective function below:
\begin{equation}
	\begin{split}
		\operatorname*{arg\,min}_{s, R, t}  \|X-(sRY+t)\|_2
	\end{split}
	\label{formular} 
\end{equation}
The above transformation matrix has seven-freedom parameters: three angles for rotation matrix, three for translation and one for scale.

Finding the solution of Equation (\ref{formular}) is difficult because of significant bad impact of cross-source challenges. 
Specifically, the impacts of cross-source challenges on the problem formulation (Eq. \ref{formular}) could be discussed from two aspects: correspondence-based methods and correspondence-free methods.

{  
\begin{itemize}
	\item optimization-based methods: the outliers, density difference, partial overlap will result in correspondence outliers. This means the found correspondences may contain large amount of inaccurate point-point correspondences. Finding the optimal solution from large correspondence outliers is difficult. Moreover, the cross-source challenges will impact the model conversion accuracy (e.g., GMM). The inaccurate models will also result in inaccurate model-based registration from data aspects.
	\item learning-based methods: the cross-source challenges will impact the feature extraction. For example, the noise and scale variation will result in different local structure, which reflect in variant extracted features. Without robust features and models, the transformation estimation is difficult.
\end{itemize}
}

\section{Research Progress}
\label{progress}
\begin{figure*}[ht]
	\centering
	\includegraphics[width=0.9\linewidth]{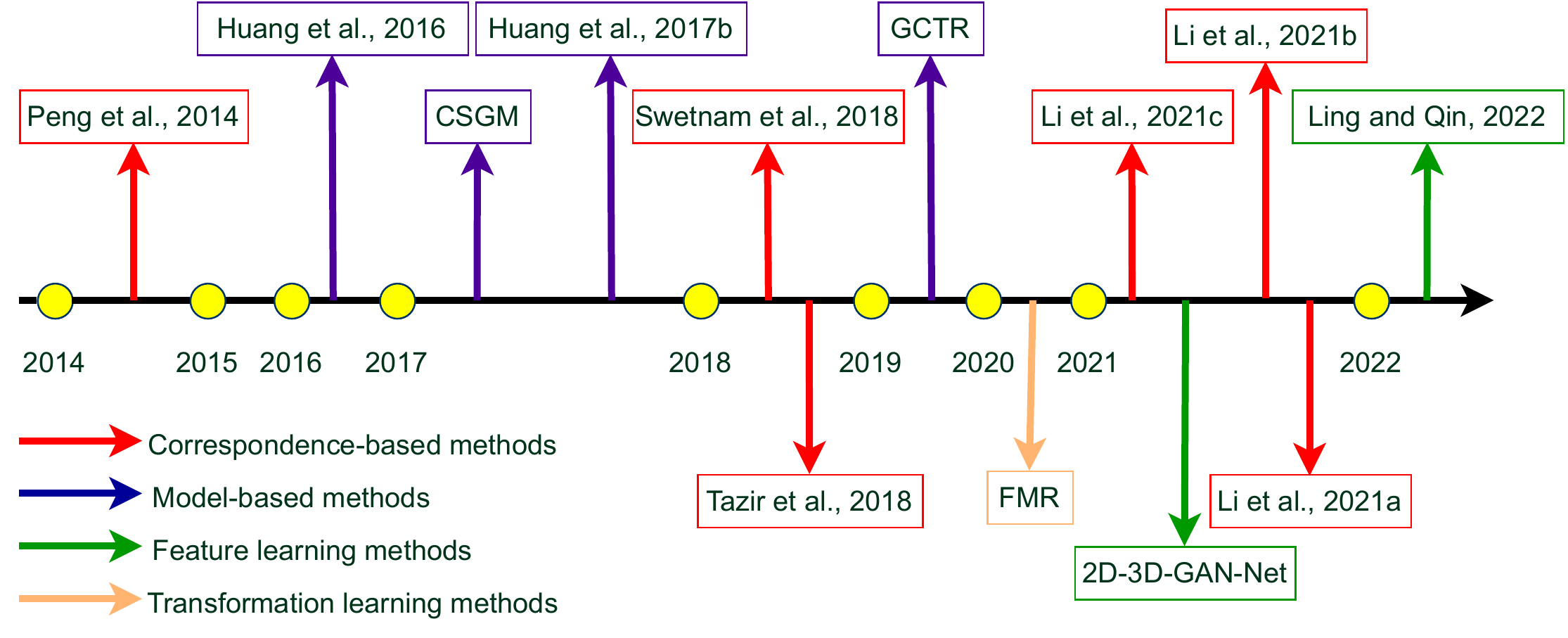}
	\caption{Chronological overview of the  CSPC registration methods.}
	\label{years}
\end{figure*}

Based on their key contributions, we divide the existing CSPC registration methods into two categories: conventional optimization methods and deep neural network methods. 
Figure \ref{category} visually shows the categorization details and figure \ref{years} illustrates the chronological overview of the development in CSPC registration. 

Since the research of the CSPC registration is just the beginning and the value of this research has not been clearly exposed to the community, we will deeply analyze the key literature in this section with the objective to help the researchers to understand the value of CSPC registration.

\subsection{Conventional optimization methods}
The main idea of conventional optimization-based methods is to design robust optimization strategies to solve the equation (\ref{formular}). Based on the data processing types, there are two kinds of algorithms: direct correspondence methods and model-based methods. This section will review details of these methods.

\subsubsection{Direct correspondence methods}
The direct methods develop optimization strategies or hand-craft features on the original point clouds directly to find the correspondences, and then  {  estimate } the transformation matrix based on these correspondences. The critical research question is how to find the accurate correspondence in the point clouds under the cross-source challenges.   {  We will review and analyze the development of this category in detail.}

\cite{peng2014street} proposes a CSPC registration method to align a small structure from motion (SFM) point cloud with a large street-view { LiDAR} point clouds. The key contributions are several pre-processing strategies and the registration part utilizes ICP. Specifically, many multi-scaled regions are cropped in the large { LiDAR} point cloud. Then, the global ESF descriptors are extracted for these cropped point clouds and SFM point cloud. After that, the top 10 matched cropped regions are selected for the SFM point clouds. Finally, the iterative closest point (ICP) algorithm is applied to further refine the top-10 matched relationships and update the order based on the refined registration error. The final top registered region is the best aligned region for SFM point cloud in the { LiDAR} point cloud. This method is efficient to solve the localization problem among cross-source point clouds.

\cite{swetnam2018considerations} utilizes the ground control point (GCP) based point cloud registration method to  merge point clouds from two 3D remote sensing techniques ({ LiDAR} and { SFM}). They found that combining point cloud data from two or more platforms allowed for more accurate measurement of vegetation than any single technique alone. This method merges the advantages of multiple sensors to benefit the agricultural application.

\cite{tazir2018cicp} proposes a surface matching optimization to utilize CSPCs on the map construction.  Specifically, a novel approach is proposed to  {  surpass} the notion of density by matching points representing each local surface of source cloud with the points representing the corresponding local surfaces in the target cloud. This method demonstrates the CSPC registration benefit for map reconstruction in robotics.

\cite{li2021linear}  proposes a hand-craft features and an optimization strategy to register MLC and photogrammetric cross-source point clouds.   Specifically, the linear features are extracted to eliminate the noise in these cross-source point clouds and a 2D incremental registration strategy is proposed  {  to simplify} the complex registration process. The experiments show that the CSPC registration method merges the merits of multiple sensors to largely contribute the photogrammetric application.

\cite{li2021laplacian} proposes a CSPC registration method to solve the occlusion problems in terrestrial laser scanning point clouds. This method will enhance the details by combining two kinds of sensors. A multi-view-projection-based vacancy filling strategy is leveraged to repair the integrity of the details.  Firstly, several pre-processing steps are applied to extract the corresponding complementary point clouds from different sources. Then, the Laplace differential coordinates are exploited to align these cross-source point clouds. The experiments show that CSPC registration achieves impressive repairing completeness in the surveys field.

OneSac \cite{li2021point} is a novel optimization method by utilizing sample consensus strategy which shows ability to overcome the cross-source challenges. Firstly, the full seven-parameter registration problem is decomposed into three subproblems,  {  which are rotation}, translation and scale estimations. Secondly, a one-point random sample consensus (RANSAC)  {  algorithm is proposed to estimate the translation} and scale parameters. Thirdly, a scale-annealing biweight estimator is proposed to estimate the rotation with the prior of the estimated scale and translation. Their experiments show that the proposed algorithm works well in cross-source point clouds.

\subsubsection{Model-based methods}
Because of large variations in the original cross-source point clouds, another category is to transform the original point cloud registration into model alignment problem. Typical examples are GMM and graph.  {  We will detailedly review and analyze the development of this category below.}

Huang et al.  \cite{huang2016coarse} proposes a CSPC registration method by utilizing the statistic property of cross-source point clouds to overcome the cross-source challenges. Specifically, a gaussian mixture models (GMM) method is applied to replace the ICP in \cite{peng2014street}. The method converted the registration problem into a GMM recovering problem. The experiments demonstrate that the statistic model obtains better robustness than directly utilizing the original points. 
However, this method use downsample to handle density difference and assume the containing ball is the same to normalize the scale difference. These two pre-processing steps are difficult to generalize to large-scale partial overlapped point clouds. How to develop a robust and general pre-processing strategy is a still a research problem.

Huang et al. \cite{huang2017coarse} further developed this GMM-based method \cite{huang2016coarse} by integrating the scale estimation into the GMM recovering process. Specifically, a scaled GMM is proposed to estimate affine transformation matrix between two point clouds. They can automatically recover the scale difference between $[0.5, 2]$. Their experiments show that the GMM-based method obtain both fast speed and accuracy in localizing the regions and estimating the pose. However, downsample pre-processing step is still required. The memory consumption will be extremely huge when the point number become large without this pre-processing step. How to reduce the memory consumption requires further work? Recent works \cite{yuan2020deepgmr,huang2022unsupervised} combined GMM with deep learning, this research direction may overcome this issue.

Apart from the GMM-based approaches, CSGM \cite{huang2017systematic} is a CSPC method to convert the registration problem into the graph matching problem to overcome the cross-source challenges. Specifically, a supervoxel method is applied to segment the cross-source point clouds into many super voxels. The center points and the relations between these these supervoxels are regarded as nodes and edges to construct two graphs. Then, an improved graph matching algorithm is proposed to find the node correspondence by considering the neighbors' coherent correspondences. Finally, the ICP is applied to further refine the registration result. The limitations of this method are that the segmentation is required to solve the density difference and only two points are considered to constraint the graph node correspondence search.

To solve the above limitation, GCTR \cite{huang2019fast} proposes a CSPC registration method to consider more neighbour correspondence constraint and convert the registration problem into a high-order graph matching problem. Specifically, triplet correspondences are extracted as the potential correspondence space and integrate them into tensors. Then, a power iteration algorithm is proposed to solve the correspondence between these two point clouds. Because this method consider more strict constraint, the experiments show more robust registration than previous graph matching-based CSPC registration algorithm. However, the segmentation will cost large amount of time and the performance is highly related to the hyperparameter (number of segments).

Recently, \cite{ling2022graph} proposes a registration method to align the over-view point clouds generated from stereo/multi-stereo satellite images and the street-view point clouds generated from monocular video images. The proposed CSPC registration algorithm segments the buildings as graph nodes and construct two graphs for the satellite-based and street-view based point clouds. Then,  the registration problem is converted into a graph-matching problem so that the topological relations between these segments is utilized to align these two point clouds. Finally, based on the matched graph nodes, a constrained bundle adjustment on the street-view image to keep 2D-3D consistencies, which yields well-registered street-view images and point clouds to the satellite point clouds. The limitation is that the accuracy is relied on the accuracy of building segmentation.

\subsubsection{Summary and trends}
The above optimization-based methods are the main branch of the current existing CSPC registration solutions. The existing methods focus on optimization strategies design, such as the graph matching and GMM optimization. Based on the above literature analysis, we found that most of the existing methods rely on pre-processing steps to reduce the bad impact of cross-source challenges to the similar level of same-source. Then, the same-source optimization strategies or hand-craft features are applied to solve the registration problem. The current existing optimization-based methods can align some example of cross-source point clouds but have not been widely demonstrated in the experiments. The robustness and accuracy are still a research question. 

The advantages of these methods are two folds: firstly, the objective function is clear to the problem and the generalization ability is predictable. Secondly, there is no need of training samples so that the deployment is easy. The limitations of the optimization-based methods are obvious from two aspects: firstly, the robustness is still an issue. The existing methods require many pre-processing steps to reduce the impact of cross-source challenges. Then, point cloud registration algorithm is designed based on the processed point clouds. The accuracy is largely relied on the quality of pre-processing. Secondly, the efficiency is still an issue when the point number increase while the point cloud usually contains large amount of points. 

The state-of-the-art CSPC registration methods are oneSac \cite{li2021point} and GCTR \cite{huang2019fast} which are sample consensus method and high-order graph matching-based algorithm. However, the current state-of-the-art algorithm is still far from the real applications because of both the efficiency and accuracy require more advanced research work. Current efficiency is guaranteed by downsampling to less points. This will impact the registration accuracy.

The research trend is to develop mathematical framework in solving the registration with more robustness. Moreover, a robust strategy to improve the efficiency without loss of accuracy is required. Last but not least, robust pre-processing strategies design is also a promising way to solve the CSPC registration.

\subsection{Deep neural network methods}
The main idea of deep neural network methods is to leverage the deep neural network to extract the feature of cross-source point clouds. Then, the transformation matrix is estimated based on feature-based correspondences or directly regression from the features. There are two kinds of learning-based methods. The first is to learn discriminative descriptor for each point. The second is to directly learn the transformation matrix.

\subsubsection{Feature learning methods}
This category aims to design neural network to extract robust point descriptors.

Recently, there are many point descriptors available, e.g., FCGF \cite{choy2019fully}, { D3Feat} \cite{bai2020d3feat} and SpinNet \cite{ao2021spinnet}. However, these point descriptors are all face challenges in cross-source point clouds.  For example, FCGF and SpinNet require a voxelization pre-processing step. This step needs to specify voxel size. However, the voxel size is difficult to specific the same as there is a scale difference in the cross-source point clouds. { D3Feat} requires k nearest neighbor to build the feature. However, this descriptor will fail when large density difference occur.

{ Apart from the aforementioned point descriptors, several methods focus on feature matching. Deep global registration (DGR) \cite{choy2020deep} designs a UNet architecture to classify whether a point pair is a correspondence or not. If the pair is correct correspondence, we recognize them as an inlier, outlier otherwise. This pipeline converts the feature matching problem as a binary classification problem(inlier/outlier).  Following this pipeline, PointDSC \cite{bai2021pointdsc} use two point pairs to consider second-order constraint and OKHF\cite{huang2022robust} considers the third-order constraint. Moreover, recent methods RPMNet \cite{yew2020rpm} and IDAM\cite{li2020iterative} show promising registration results on partial point clouds. However, detailed discussion these methods beyond the scope of this paper, please refer to same-source point cloud registration survey for more information \cite{huang2021comprehensive}. }

Liu et al. \cite{liu2021matching} propose a network, 2D3D-GAN-Net, to learn the local invariant cross-domain feature descriptors of 2D image patches and 3D point cloud volumes. Then, the learned local invariant cross-domain feature descriptors are used for matching 2D images and 3D point clouds. The Generative Adversarial Networks (GAN) is embedded into the { 2D3D-GAN-Net}, which is used to distinguish the source of the learned feature descriptors, facilitating the extraction of invariant local cross-domain feature descriptors. Experiments show that the local cross-domain feature descriptors learned by 2D3D-GAN-Net are robust, and can be used for cross-dimensional retrieval on the 2D image patches and 3D point cloud volumes dataset. In addition, the learned 3D feature descriptors are used to register the point cloud for demonstrating the robustness of learned local cross-domain feature descriptors.

\subsubsection{Transformation learning methods}
The transformation learning methods aims to estimate the transformation directly using the neural networks.

There is not specific transformation learning method for cross-source point cloud registration. Based on our survey, FMR \cite{huang2020feature} proposes a feature-metric registration method to align two point clouds by minimizing their feature-metric projection error. This is the current only method shows potential in solving CSPC registration problem by directly estimating the transformation. Specifically, FMR uses a PointNet (without T-Net) to extract a global feature of two point clouds. Then, the feature-metric projection error is calculated as the feature difference of two point clouds. After that, the Lukas-Kanadle (LK) algorithm is applied to estimate the transformation increments. This process iteratively runs for 10 times to obtain the final transformation.

\subsubsection{Summary and trends}
Although there are several initial trials, both the accuracy and robustness require significantly improvement for CSPC registration. The research of using neural networks to solve the CSPC registration problem is almost a blank space. This branch of research has been overlooked. The reason maybe the current open dataset is all same-source and the value of CSPC registration has not been clearly exposed to the community. 

The current state-of-the-art algorithm is 2D3D-GAN-Net. The advantages of the method are fast and relatively accurate. However, the limitations lie in the generalization performance on unseen scenes.

The trend could be three aspects. Firstly, there is no large-scale cross-source dataset available. This is the big stone to obstacle the neural network development in this area. A large-scale cross-source dataset is an urgent research direction. Secondly, a point descriptor that is robust to cross-source challenges is also high required. How to leverage multimodal information \cite{huang2022imfnet,huang2022gmf} or pretrained models \cite{huang2022clip2point,huang2022frozen} to improve the ability of point descriptor. Thirdly, an end-to-end learning framework for CSPC registration is also required in terms of both efficiency and accuracy requirement.

\subsection{Comparison of Conventional and Learning methods}
The existing conventional methods require several pre-processing steps to remove the outliers, density and scale differences. These steps are usually time-consuming and require a lot of manual work. Both the robustness and accuracy need further research work to improve. In contrast, the learning-based methods require few pre-processing steps and the efficiency is high. However, the learning-based methods are only at the start point. Many research works are required to facilitate the development in this field.

\subsection{Performance evaluation and analysis}
{ The current existing cross-source point cloud registration methods are evaluated at different setting and benchmarks. To better evaluate the performance and investigate the development of this field, we evaluate the current cross-source point cloud registration on a cross-source benchmark \cite{huang2021comprehensive}.  Because the code of many existing cross-source point cloud registration methods have not been released, we only compare the key components that are open-sourced in their methods. {\cite{peng2014street},\cite{swetnam2018considerations}, \cite{tazir2018cicp}, \cite{li2021linear},} \cite{li2021laplacian}, \cite{li2021point}, \cite{huang2016coarse}  use ICP to register the cross-source point clouds. We select the ICP that implemented in Open3D. OneSac\cite{li2021point}  is an improved RANSAC, we have tried to implement it but fail to obtain comparable results. Instead, we choose RANSAC that is implemented in Open3D. We also compare with GCTR \cite{huang2019fast}, GMM-based method \cite{huang2016coarse} and FMR \cite{huang2020feature}. In addition, we have included RPMNet\cite{yew2020rpm} and IDAM\cite{li2020iterative} in our benchmark since they also show ability in aligning cross-source point clouds. We also evaluate the feature-matching methods including learning-based (e.g., DGR \cite{choy2020deep}) and optimization-based (e.g., FGR \cite{zhou2016fast}) methods. We uses the evaluation metric in DGR to evaluate their performance on 3DCSR dataset and the results are shown in Table \ref{t2}. 

Table \ref{t2} shows that the current cross-source point cloud registration methods face difficulty handling cross-source point clouds with a wide variety. The generalization is a big remaining problem to be addressed. The DGR achieves the current highest generalization accuracy, which is only 36.6\%. In this new GPT era, developing high-generalization registration algorithms is an urgent research problem that could benefit broad audiences in many applications.
}

\begin{table}[ht]
	\centering
	\begin{tabularx}{0.9\linewidth}{l|X|X|X|X}\hline
		Method  &  RR  & TE    &  RE    & Time      \\ \hline
		ICP     & 24.3 & 0.38  &  5.71  & 0.19      \\
		RANSAC  & 3.47 & 0.13  &  8.30  & 0.03      \\
		GCTR    & 0.5  & 0.17  &  7.46  & 15.8      \\
		GM-CSPC & 1.0  & 0.71  &  8.57  & 18.1      \\
		FMR     & 17.8 & 0.10  &  4.66  & 0.28      \\
		FGR     & 1.49 & 0.07  &  10.74 & 2.23      \\
		DGR     & 36.6 & 0.04  &  4.26  & 0.87      \\ 
		RPMNet  & 27.7 & 0.13  &  10.04 & 2.16      \\ 
		IDAM    & 1.98 & 0.13  &  11.37 & 0.48      \\ 
		\hline
		
	\end{tabularx}
	\caption{Performance evaluation on cross-source point cloud registration benchmark 3DCSR.}
	\label{t2}
\end{table} 

\section{Open Research Directions}
Based on the above literature analysis, there are several research directions for the CSPC registration problem.

\subsection{Benchmark of CSPC registration}
Based on the above survey of Section \ref{progress}, we find that the evaluation is not unified. Different methods evaluate using their own data and metrics. It is urgent to propose a benchmark to fairly evaluate the CSPC registration and promote the algorithm development in this field.

Possible solution is to leverage the RGB images and { LiDAR} data in the outdoor database to build a cross-source dataset. Another way could use two kinds of 3D sensors (e.g., Intel Realsense and { LiDAR}) to capture a large-scale dataset to build the benchmark.

\subsection{Robust pre-processing strategies}
Cross-source challenges contain large variations in density, outliers and scale difference. These large variations are the key cues to make the existing state-of-the-art same-source point cloud registration algorithms fail. Based on the above literature analysis, most of the existing optimization-based methods utilize pre-processing strategies to make the original cross-source point clouds be similar to same sources. { Then, the well-developed same-source point cloud registration methods \cite{zhou2016fast,yang2020teaser,wu2022evolutionary,wu2022inenet,wu2022multi} can work on these processed data. For example, if the pre-processing strategy can handle the density and scale variants, the well-known same-source method TEASER \cite{yang2020teaser} can work well on those pre-processed point clouds. Moreover, the recent method \cite{wu2022multi}, a milestone work to use evolutionary theory for multiview point cloud registration, can play a role in aligning multiview point clouds. However, the current pre-processing strategies still need to be improved for general cases.}

Therefore, one research direction is the pre-processing strategies development, which aims to reduce the bad impact of density, outliers and scale variations. The difficulty is that these variations are significant and usually mixture. If the pre-processing step could do well, we can leverage the existing same-source point cloud registration algorithms.

\subsection{Learning-based algorithms}
Leveraging the performance of neural networks ia another research direction to solve the CSPC registration problem. There are two options to develop learning-based methods. 

Firstly, a robust learning-based descriptor is urgent for cross-source point clouds. Deep learning largely force the performance improvement in same-source point cloud registration \cite{choy2019fully,bai2020d3feat}. The development learning-based descriptor in CSPC registration will be very promising in improving both the accuracy and efficiency. The challenges are how to extract the descriptor by overcoming the density and scale difference simultaneously.  

Secondly, an end-to-end learning-based method is another research direction in leveraging deep learning. The existing research in same-source point cloud registration \cite{choy2020deep} shows that the end-to-end learning-based methods will achieve both high accuracy and efficiency.

\section{Applications}
\label{applications}
Because CSPC registration is a technology that could utilize the advantages of several types of sensors, it has a wide range of applications.  This section introduces the role in several applications fields. This will provide insights on how to use the research technology in real applications and why CSPC registration is an important research question from application aspects.

\subsection{Robotics}
In the robotics field, the 3D point clouds are widely utilized \cite{tazir2018cicp}. Because of { LiDAR} point cloud has the merits of fast and accurate, this sensor is usually used to generate map, which provides basic environment information. In the applications, the cheaper sensors, e.g., Intel realSense, are utilized to obtain detailed point clouds. Registration between these two types of sensor data is able to build detailed map or provide consumer-affordable service such as localization. Recently, autonomous driving is an active branch of robotics which utilizes multiple sensors in their vision system. CSPC registration provides a solution to utilize the 3D data in the autonomous driving field. For example, detailed 3D map reconstruction and accurate visual-based localization in GPS-denied regions. Among these applications, accurate, fast and efficient CSPC registration is the key. Research in this area is timely and has high value for field robotics.

\subsection{Remote sensing}
3D point cloud has a high value in the remote sensing regarding the fields of forest \cite{li2021linear}, agricultural \cite{swetnam2018considerations} and survey \cite{li2021laplacian}. These applications, such as remotely sensing recent growth, herbivory, or disturbance of herbaceous and woody vegetation in dryland ecosystems, requires high spatial resolution and multi-temporal depth point clouds. Three dimensional (3D) remote sensing technologies like { LiDAR}, and techniques like { SFM} photogrammetry, each have strengths and weaknesses at detecting vegetation volume and extent, given the instrument's ground sample distance and ease of acquisition. Yet, a combination of platforms and techniques might provide solutions that overcome the weakness of a single platform. Combining point cloud data and derivatives (i.e., meshes and rasters) from two or more platforms allowed for more accurate measurement of ground elements (e.g., height and canopy cover of herbaceous and woody vegetation) than any single technique alone.  Therefore, development of CSPC registration has a high value in the remote sensing field.

\subsection{Construction}
BIM (Building Information Modelling) is a new way for the generation of information storage and manipulation systems that is widely used for construction purposes and building management. Previous computer-aided BIM designs were limited to simple guides and theoretical planning since there is no requirement of any interaction with the real physical world.
Recently, point cloud can overcome this limitation and offer the ability to align the digital models with the physical space in great detail. The reason is that point cloud provides the ability to effectively import 3D physical space into a digital format and augment existing digital models. 

Comparing the real-time { LiDAR} point clouds and BIM can generate valuable information. For example, the building quality check. Developing a fast and high accurate CSPC registration algorithm with construction field knowledge is timely and will contribute to the construction field.

\section{Conclusion}
The CSPC registration is an emerging research topic in the 3D field that appeared from the fast development of sensor technologies. It can combine the advantages of several sensors and overcome the limitation of single sensor. The CSPC registration will force the development of computer vision tasks like 3D reconstruction and visual-based localization in a new way, and significantly contribute to many application fields. There are still a lot of remaining research questions to be solved. We hope that this survey paper provides an  overview  of the  challenges  and  the 
recent progress and some future directions as well as the applications fields in CSPS registration to the 3D computer vision community.

\bibliographystyle{elsarticle-num}
\bibliography{ijcai22}

\end{document}